\documentclass[journal]{IEEEtran}
\usepackage{xcolor}
\usepackage{booktabs}
\usepackage{epsfig}
\usepackage{graphicx}
\usepackage{amsmath}

\usepackage{amssymb}

\usepackage{verbatim}

\usepackage{url}
\usepackage{tabularx}
\usepackage{multirow}
\usepackage{subfigure}
\usepackage{amsmath,dsfont}
\usepackage{array}
\usepackage{enumitem}

\usepackage{pifont}
\usepackage{threeparttable}
\usepackage{bbding}

\usepackage{verbatim}

\usepackage{colortbl}
\definecolor{mygray}{gray}{.75}

\usepackage{xcolor}

\usepackage[normalem]{ulem}

\usepackage{makecell}
\usepackage{multirow}
\usepackage[linesnumbered,ruled,vlined]{algorithm2e}
\usepackage{algorithmic}

\begin{document}
%

\title{Recurrent Aligned Network for Generalized Pedestrian Trajectory Prediction}
%
%
%

	\author{Yonghao Dong,
	Le~Wang,~\IEEEmembership{Senior Member,~IEEE,}
	Sanping~Zhou,~\IEEEmembership{Member,~IEEE,}
	Gang~Hua,~\IEEEmembership{Fellow,~IEEE,}
	and~Changyin Sun,~\IEEEmembership{Senior Member,~IEEE}
	\thanks{
 
 This work was supported in part by the National Key R\&D Program of China under Grant 2021YFB1714700, in part by NSFC under Grants 62088102, 62106192, and 12326608, in part by Natural Science Foundation of Shaanxi Province under Grant 2022JC-41, and in part by Fundamental Research Funds for the Central Universities under Grant XTR042021005. \textit{(Corresponding author: Le Wang.)}}%
	\thanks{Yonghao Dong, Le Wang and Sanping Zhou are with the National Key Laboratory of Human-Machine Hybrid Augmented Intelligence, National Engineering Research Center for Visual Information and Applications, and Institute of Artificial Intelligence and Robotics, Xi'an Jiaotong University, Xi'an, Shaanxi 710049, China. (e-mail: yhdong@stu.xjtu.edu.cn; \{lewang, spzhou\}@mail.xjtu.edu.cn)}
	\thanks{Gang Hua is with the Multimodal Experiences Lab, Dolby Laboratory, Bellevue, WA 98004, USA. (e-mail: ganghua@gmail.com)}
	\thanks{Changyin Sun is with the School of Artificial Intelligence, Anhui University, Hefei, Anhui 230039, China. (e-mail: cysun@ahu.edu.cn)}}

\markboth{IEEE Transactions on Circuits and Systems for Video Technology,~Vol.~x, No.~x}
{Shell \MakeLowercase{\textit{et al.}}: Bare Demo of IEEEtran.cls for IEEE Journals}
%



\maketitle

\begin{abstract}
 Pedestrian trajectory prediction is a crucial component in computer vision and robotics, but remains challenging due to the domain shift problem. 
 Previous studies have tried to tackle this problem by leveraging a portion of trajectory data from the target domain to fine-tune the model. 
 However, such domain adaptation methods are impractical in real-world scenarios, as it is infeasible to collect trajectory data from all potential target domains. 
 In this paper, we study a new task named generalized pedestrian trajectory prediction, with the aim of generalizing the model to unseen domains without accessing their trajectories. 
 To tackle this task, we further introduce a Recurrent Aligned Network~(RAN) to minimize the domain gap through domain alignment. 
 Specifically, we devise a recurrent alignment module to effectively align the trajectory feature spaces at both time-state and time-sequence levels by the recurrent alignment strategy.
 Furthermore, we introduce a pre-aligned representation module to combine social interactions with the recurrent alignment strategy, which aims to consider social interactions during the alignment process instead of just target trajectories. We extensively evaluate our method and compare it with state-of-the-art methods on three widely used benchmarks. The experimental results demonstrate the superior generalization capability of our method. 
 Our work not only fills the gap in the generalization setting for practical pedestrian trajectory prediction, but also sets strong baselines in this field.
\end{abstract}


%
\IEEEpeerreviewmaketitle

\section{Introduction}\label{sec:introduction}
\begin{figure*}[t]
	\centering
	\includegraphics[width=1\linewidth]{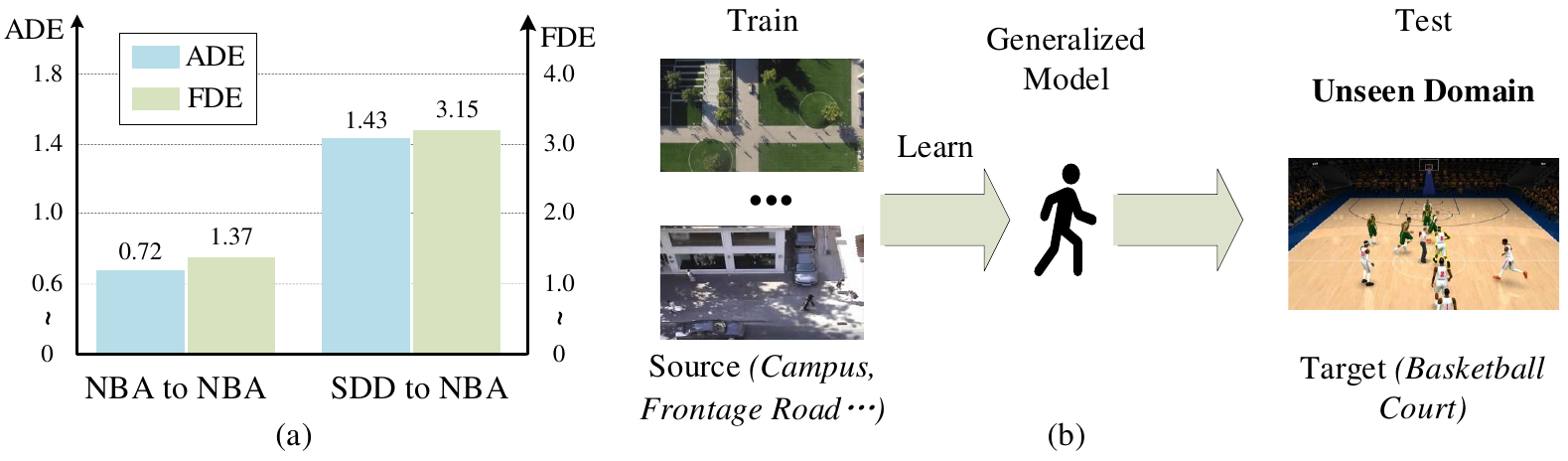}
	\caption{	(a) The graph shows performance decrease for SocialVAE~\cite{xu2022socialvae}, which demonstrates current work can't address domain shift challenge well for pedestrian trajectory prediction. `A to B' below X-axis denotes that the model is trained in source dataset `A' and tested in target dataset `B'.
		(b) Definition of the generalized pedestrian trajectory prediction. 
		In this generalization setting, the model is trained in multiple source domain datasets and tested in the target domain dataset.
	}
	\label{fig1}
\end{figure*}

Pedestrian trajectory prediction aims to estimate the future movements of pedestrians given their observed trajectories. This task is crucial in various applications such as autonomous driving~\cite{li2024beyond,zhou2024drivinggaussian,tonderski2024neurad}, visual recognition~\cite{hou2024conv2former, du2024probabilistic, Lee_2023_CVPR}, navigation~\cite{an2024etpnav, Kamath_2023_CVPR, Sixian_2023_CVPR}, and video surveillance~\cite{chen2024top, liu2024human, chena2024local}.
Despite significant progress in this field, pedestrian trajectory prediction remains challenging due to the domain shift problem.
This problem arises when a model encounters scenes significantly different from its training dataset, resulting in performance degradation.
For example, the motion styles of students on a university campus differ significantly from those of players on a basketball court. Consequently, a predictor trained on campus datasets (\textit{e.g.}, SDD) may excel in that scene, but encounter difficulties with basketball court datasets (\textit{e.g.}, NBA), as depicted in Figure~\ref{fig1}(a). 
Recent studies~\cite{xu2022adaptive, ivanovic2023expanding} have identified the domain shift problem in pedestrian trajectory prediction and mitigated it through domain adaptation. 
However, in real-world scenarios like autonomous driving, domain changing is inevitable. Collecting trajectory data for all possible target domains is impractical, making the application of domain adaptation using data from target domains for model fine-tuning unfeasible in practice.
Therefore, researching pedestrian trajectory prediction in more general scenarios, specifically under the domain generalization setting, is crucial for applications like autonomous driving.

In this paper, we tackle the challenge of predicting pedestrian trajectories in an unseen target domain dataset, without requiring data from the target domain during model training, as illustrated in Figure~\ref{fig1}(b).
We term this task as \textit{Generalized Pedestrian Trajectory Prediction}. 
Generalized pedestrian trajectory prediction is more challenging and meaningful compared to previous settings. 
Specifically, unlike conventional settings that do not account for the domain shift problem, our approach is more suitable for the dynamically changing environments of autonomous driving. Compared to the previous domain adaptation setting, our method better addresses the difficulty of data collection in unknown scenarios.

One common and effective method to address this challenge is through domain alignment, which minimizes the data distribution gap between the source and target domains in high-dimensional feature spaces by using alignment losses such as Maximum Mean Discrepancy (MMD)~\cite{li2018deep}, Correlation Alignment Distance (CORAL)~\cite{sun2016deep}, and adversarial loss~\cite{wu2020dual}.
\textcolor{black}{Nevertheless, traditional domain alignment strategies are not well suited for the pedestrian trajectory prediction task because 1) they overlook the sequential property of the pedestrian trajectory prediction task and align different domains directly, without considering the alignment at both time-state and time-sequence levels within trajectories;} 2) they overlook the surrounding alignment, ignoring that the predicted trajectory is not only associated with that specific trajectory but also related to other interconnected trajectories within the scene.

To address these issues, we further introduce a \textit{Recurrent Aligned Network (RAN)}, a new alignment framework tailored for generalized pedestrian trajectory prediction, as illustrated in Figure~\ref{fig3}.
RAN deals with domain alignment by considering both time-state and time-sequence alignments, along with the social interaction dynamics inherent in pedestrian trajectories.
This advancement is facilitated by two key modules: a pre-aligned representation module and a recurrent alignment module.
The recurrent alignment module aims to minimize the domain gap at both time-state and time-sequence levels by a novel recurrent alignment strategy.
Specifically, we first input trajectory representations from different domains at the starting time step into the GRU units and update the network using alignment losses calculated by the output latent variables.
Then, we input the trajectory representations of the next time step and the previously aligned latent variables into GRU and update the network with the new alignment losses. This iterative process continues until all time steps' trajectory representations have been processed, enabling the network to grasp generalized trajectory features across different domains, thereby aligning the gap between domains.
Furthermore, the pre-aligned representation module focuses on accounting for the human interactions at each time step for more comprehensive alignment, achieved through stepwise attention mechanisms. 
Specifically, social interactions along with the target trajectory features are regarded as trajectory representations at each time step, which are used for the recurrent alignment process.
Finally, in addition to these two key modules, we adopt an RNN-based trajectory decoder to convert generalized trajectory features into multiple possible future trajectory predictions.

The main contributions of our work are summarized as three-fold:
\begin{itemize}
	\item \textcolor{black}{We identify the domain shift problem and focus on a generalized approach for the pedestrian trajectory prediction task. Furthermore, we emphasize the differences in domain generalization between trajectory prediction and most other tasks, highlighting that domain generalization in trajectory prediction must consider its temporal and recurrent nature.}
	\item We propose a Recurrent Aligned Neural Network to enhance the model's generalization capability for generalized pedestrian trajectory prediction.
	\item We present a recurrent alignment module to minimize the domain gap at both time-state and time-sequence levels. Furthermore, we introduce a pre-aligned representation module, which considers human social interactions instead of solely target trajectories, into the domain alignment process.
	\item Extensive experiments on three widely used benchmarks, including ETH-UCY, SDD, and NBA datasets, show that our method outperforms all existing methods and demonstrates our superior generalization capabilities.
\end{itemize}

\section{Related Work} \label{sec:rel-work}

\begin{table*}[h]
	\caption{Examples of domain gaps in different attributes between ETH-UCY, SDD and NBA datasets.}
	\centering
	\resizebox{1\linewidth}{!}{
		\setlength{\tabcolsep}{1.5em}%
		\begin{tabular}{c|cccc}
			\toprule
			Dataset 		& Main Scenario Type  &	Agent Type	& \# of Pedestrian &	\# of Trajectory	\\
			\midrule
			ETH-UCY 	&	Outdoor pedestrian paths		&	Pedestrian																&	1536	&	37270 	\\
			SDD 		&	University campus				&	\makecell{Bicyclist, Pedestrian,\\ Skateboarder, Cart, Car, Bus}		&	 5232	&	11323\\
			NBA 		&	Basketball courts				&	Basketball player														& 	3155	&	257230\\
			
			\bottomrule
		\end{tabular}
	}
	\label{table_GAP}
\end{table*}

\subsection{Pedestrian Trajectory Prediction}
The objective of pedestrian trajectory prediction~\cite{peng2023mrgtraj,sun2021reciprocal,berenguer2020context} is to anticipate future movements of pedestrians based on the observed trajectories and their surrounding context.  This task faces several challenges, including human interactions~\cite{shi2021sgcn, li2022graph, bae2022learning}, multimodality~\cite{shi2022social, xu2022socialvae, chen2022multimodal}, long-tail distribution~\cite{wang2023fend}, and momentary prediction~\cite{sun2022human}.

Taking advantage of advancements in deep learning, pedestrian trajectory prediction has made significant progress in addressing these challenges. For the challenge of human interaction, 
\textcolor{black}{DifTraj~\cite{johnson2016malmo} considers human intention features to enhance interaction modeling.}
MSRL~\cite{wu2023multi} proposes to model complex human interactions using three separate branches: the temporal branch, the spatial branch, and the joint branch.
\textcolor{black}{Furthermore, due to the close relationship between tracking and trajectory prediction tasks, some interaction modeling methods~\cite{dong2022adaptive, bergmann2019tracking, yin2020unified, liu2020multiple,li2020graph} used in tracking tasks can also provide significant insights for trajectory prediction.}
In terms of multimodality challenge, SIT~\cite{shi2022social} proposes a social interpretable tree to represent various possible future trajectories and utilizes a teacher-student network to ensure that the learned future trajectory distributions align with the tree structure. SocialVAE~\cite{xu2022socialvae} employs a conditional variational autoencoder (CVAE) to capture the distribution of multimodal future trajectories within a latent space. SICNet~\cite{dong2023sparse} proposes a method that first generates representative sparse instances for future modalities and then generates multimodal predictions conditioned on these sparse instances.
To handle the long-tail distribution problem, FEND~\cite{wang2023fend} introduces an improved contrastive learning framework to effectively identify less frequent trajectory patterns.
For momentary prediction, MOE~\cite{sun2022human} incorporates a momentary observation feature extractor, allowing efficient use of valuable momentary information in the prediction process.
Furthermore, the recent work, T-GNN~\cite{xu2022adaptive}, has identified the domain shift problem in pedestrian trajectory prediction, where a trained trajectory prediction model experiences performance degradation when transferred to unfamiliar environments. T-GNN addresses this issue using domain adaptation, where a portion of data from the target environment is used to fine-tune the model. However, this approach is not practical in trajectory prediction due to the difficulty in obtaining data from the target environment.

Thus, in this paper, we concentrate on a more practical setting, \textit{i.e.}, domain generalization, for pedestrian trajectory prediction. We also propose an efficient recurrent aligned neural network to tackle this domain shift problem under the domain generalization setting.

\subsection{RNNs for Sequence Prediction}
Recurrent Neural Networks (RNNs)~\cite{weng2022boosting, zhang2023differentiating, wang2023versatile} constitute a versatile class of dynamic models that extend the capabilities of feedforward networks, enabling sequence generation in diverse tasks such as speech recognition~\cite{saon2021advancing}, machine translation~\cite{farooq2023multi}, and image captioning~\cite{luo2023semantic}.
Pedestrian trajectory prediction is inherently a time-sequence prediction task, making it well-suited for feature extraction using RNN-based architectures like Long Short-Term Memory~(LSTM) and Gated Recurrent Unit~(GRU).
For example, PCCSNet~\cite{Sun_2021_ICCV} adopts the LSTM as encoders to extract observed and future trajectory features for deep clustering, classification, and further synthesis, and then generate the future prediction by the obtained features from the LSTM.
Bitrap~\cite{yao2021bitrap} extends the RNN into a GRU-based bi-directional decoder to improve the accuracy of long-term trajectory prediction and reduce collision rates.
SocialVAE~\cite{xu2022socialvae} introduces a temporal variational encoder architecture for pedestrian trajectory prediction, which combines RNNs with variational autoencoders~(VAEs) at each time step. This integration incorporates multimodality into each time step, enhancing both the interpretability and accuracy of multimodal pedestrian trajectory prediction.
However, pedestrian trajectory prediction is more complex than traditional time-sequence prediction, as the constantly changing environments make model generalization crucial. Traditional RNNs lack the ability to extract generalized features when handling temporal data across different scenes. As a result, these models may perform well in-domain but struggle with out-of-domain scenarios.

In this paper, we focus on addressing the domain generalization challenge in pedestrian trajectory prediction. To achieve this, we enhance the RNN with a time-sequential alignment strategy that employs recurrent alignment losses, effectively bridging the data gap across various domains.

\begin{figure*}[t]
	\centering
	\includegraphics[width=\linewidth]{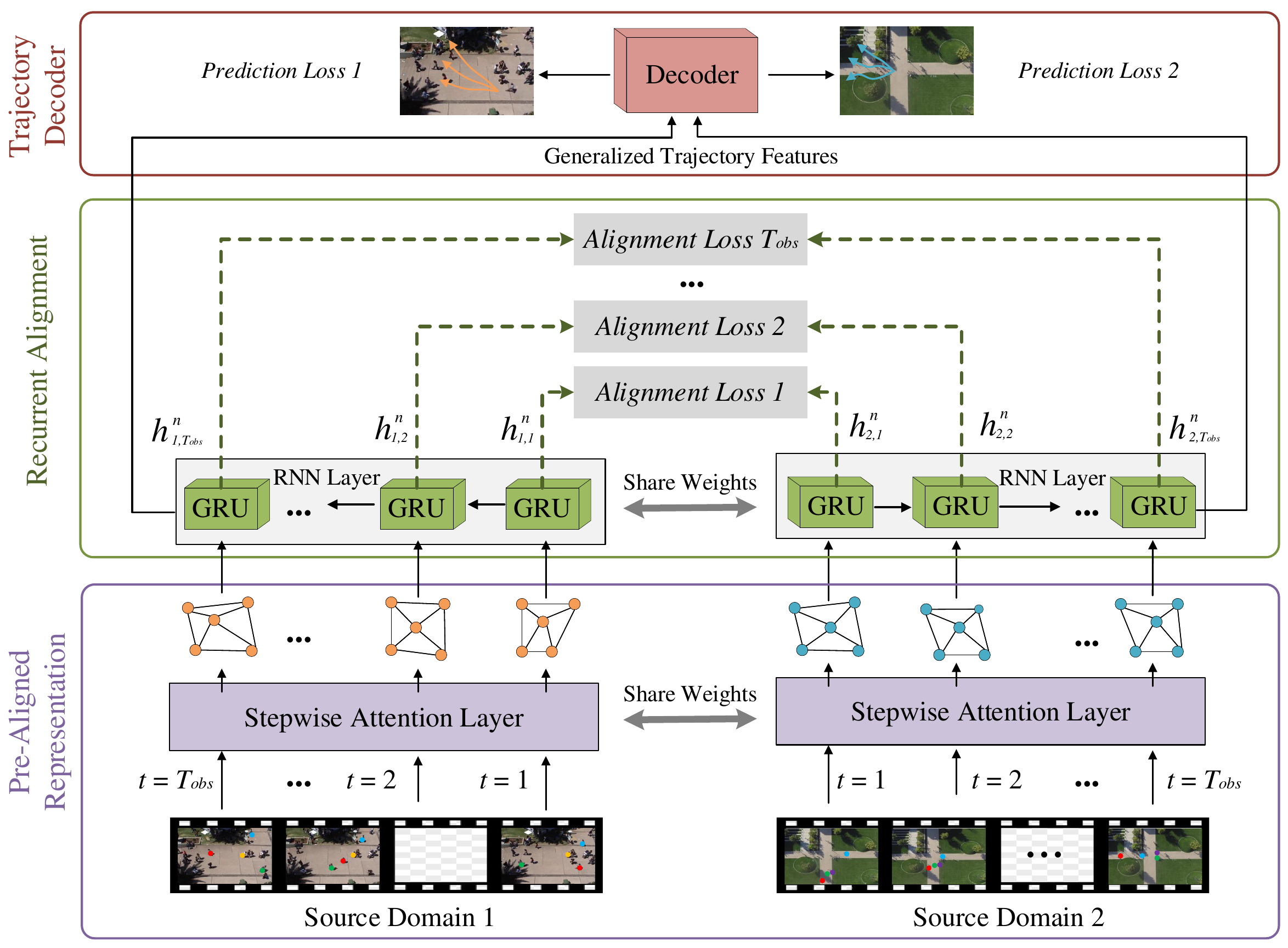}
	\caption{ 
		The framework of the training phase of RAN.
		RAN need two source domains in the training process.
		Given trajectories of two different domains, we first model the pre-aligned representation of trajectories by the stepwise attention layers. 
		Then we align the two domains' representation at both the time-state and time-sequence levels using the recurrent alignment module. 
		Finally, we decode the well-aligned features~(generalized trajectory features) into trajectory predictions.
		Note that the RNN layers and the stepwise attention layers share weights, respectively. 
	}
	\label{fig3}
\end{figure*}

\subsection{Domain Generalization}
Domain generalization~(DG)~\cite{kang2022style, zhang2022towards, wan2022meta, li2021progressive} aims to acquire representations that remain invariant across various domains, enabling the model to effectively generalize to unfamiliar domains.
Both domain generalization and domain adaptation~(DA)~\cite{zhou2023unsupervised, wang2023dynamically, kennerley20232pcnet, liu2023cot} tackle the challenges posed by domain shift.
However, the most significant difference between DA and DG is that DG does not require access to the target domain during training.
In real-world scenarios, DG is more practical and challenging than DA, mainly because collecting data from all target domains can be arduous in real-world situations.
A recent work~\cite{xu2022adaptive} demonstrates that different scenarios for pedestrian trajectory prediction have a domain shift problem, as illustrated in Table~\ref{table_GAP}, and addresses this domain shift problem within the domain adaptation setting, which may be impractical in real-world scenarios due to the difficulty of obtaining target domain data.
Moreover, Frenet~\cite{ye2023improving} solves the domain shift problem from the perspective of domain generalization. 
Nevertheless, it is important to note that this work focuses primarily on vehicle trajectory prediction, limiting its direct applicability to pedestrian trajectory prediction due to distinctions in scenarios, experimental settings, and technology frameworks.

Domain generalization methods encompass various methods such as domain alignment~\cite{li2018domain, chen2020angular}, meta-learning~\cite{wan2022meta, du2020learning}, and data augmentation~\cite{volpi2018generalizing, zhou2020learning}.
\textcolor{black}{For example, AdapTraj~\cite{qian2024adaptraj} contributes to pedestrian trajectory prediction in generalization settings by specifically focusing on decoupling domain-invariant and domain-specific features of trajectories. MetaTra~\cite{li2024metatra} solves the domain shift problem in pedestrian trajectory prediction through meta-learning. Unlike previous works, this paper places greater emphasis on the recurrent nature of trajectory data in domain generalization. Consequently, we have adopted a domain alignment strategy and enhanced it by incorporating a recurrent alignment strategy, tailoring it specifically for generalized pedestrian trajectory prediction.}



\section{Proposed Method} \label{sec:method}

The overall framework of the RAN is illustrated in Figure~\ref{fig3}. Due to the property of the domain alignment method, RAN needs two source domains in the training process for alignment.
RAN consists of three main components: 1) a pre-aligned representation module that models trajectory representations from each source domain for alignment, 2) a recurrent alignment module that aligns each trajectory domain from both the time-state and time-sequence levels, and 3) a trajectory decoder module that decodes the generalized trajectory features into multimodal future predictions.

\subsection{Problem Formulation}
The objective of pedestrian trajectory prediction is to predict the future trajectory coordinates of the agent based on its observed trajectory and neighboring trajectories. 
Mathematically, given a pedestrian in domain $i$, let $\mathbf{O}^i = [o_{i,1}, o_{i,2}, ..., o_{i,T_{obs}}]$ be the observed trajectory over $T_{obs}$ time steps, where $o_{i,t} \in \mathbb{R}^2$ denotes the 2D spatial coordinate at time step $t$ in domain $i$.
Let $\mathcal{B}^i = [B_{i,1}, B_{i,2}, ..., B_{i,T_{obs}}]$ be the neighboring trajectories of $\mathbf{O}^i$, where $\mathbf{B}_{i,t} \in \mathbb{R}^{2 \times a}$  is the 2D spatial coordinates of all neighboring trajectories at time step $t$. $a$ is the number of neighbors.
The future trajectory is $\mathbf{Y}^i = [y_{i,T_{obs}+1}, y_{i,T_{obs}+2}, ..., y_{i,T_{pred}}]$, where $y_{i,t} \in \mathbb{R}^2$ denotes the 2D spatial coordinate at time step $t$ in domain $i$. 
The goal is to train a prediction model $g(\cdot)$, so that the prediction $(\hat{\mathbf{Y}}^i)^{min}$ with minimum error in multimodal predictions $(\hat{\mathbf{Y}}^i)^K = g(\mathbf{O}^i, \mathcal{B}^i)$ is as close to the ground truth $\mathbf{Y}^i$ as possible.

Note that generalized pedestrian trajectory prediction requires training the model $g(\cdot)$ in the source domains and testing $g(\cdot)$ in different target domains.
Due to the property of the domain alignment method, we incorporate two source domains during the training phase.	

\subsection{Pre-Aligned Representation}
Pedestrian trajectory prediction is more challenging than traditional time-sequence prediction tasks, because it necessitates accounting for both the time-sequence property and the human interaction property at the time-state level.
Before modelling the time-sequence information of pedestrian trajectories, we first model the interactions between pedestrians at each time step.
To capture the trajectory interactions within a scene at each time step, we adopt a stepwise attention mechanism to model the influence of neighboring trajectories on the agent to be predicted.

Given a trajectory $\mathbf{O}^i$ in domain $i$ and its neighboring trajectories $\mathcal{B}^i$, we first extract their high-dimensional feature embeddings at each time step as follows:
\begin{equation}
	\begin{split} 
		\mathbf{F}_{i,t}^o &= f_o(o_{i,t}), \\
		\quad \mathbf{F}_{i,t}^b &= f_b(B_{i,t}),\\
	\end{split}
\end{equation} 
where $\mathbf{F}_{i,t}^o$ and $\mathbf{F}_{i,t}^b$ are the feature embeddings of $o_{i,t}$ and $B_{i,t}$, respectively.
$f_o(\cdot)$ and $f_b(\cdot)$ are nonlinear multilayer perceptron (MLP) encoders.
To consider the influence of neighboring agents, we adopt the attention mechanism to model the social interaction at each time step as follows:
\begin{equation}
	\begin{split} 
		Q_{i,t} &= \phi(\mathbf{F}_{i,t}^o, W_{i,t}^{q}), \\
		K_{i,t} &= \phi(\mathbf{F}_{i,t}^b, W_{i,t}^{k}),  \\
		V_{i,t} &= \phi(\mathbf{F}_{i,t}^b, W_{i,t}^{v}),  \\
		A_{i,t} &= \text{softmax}(\frac{Q_{i,t} (K_{i,t})^T}{\sqrt{d_k}})V_{i,t}  ,
	\end{split}
\end{equation} 
where $\phi(\cdot, \cdot)$ denotes the linear transformation. 
$W_{i,t}^{q}$, $W_{i,t}^{k}$ and $W_{i,t}^{v}$ are learnable parameters.
$\mathbf{Q}_{i,t} \in \mathbb{R}^{1 \times d}$, $\mathbf{K}_{i,t} \in \mathbb{R}^{a \times d}$, and $\mathbf{V}_{i,t} \in \mathbb{R}^{a \times d}$ are the query, key and value of the attention at time step $t$ in domain $i$, respectively.
$\mathbf{A}_{i,t} \in \mathbb{R}^{1 \times d}$ is the output of the stepwise attention.
$\sqrt{d_k}$ is a scaled factor to ensure numerical stability.
The feature of the target pedestrian at the time step $t$ should consist of its own coordinate information and its interaction with surrounding pedestrians.
Hence, we can formulate the trajectory representation at time step $t$ in domain $i$ by considering the observation $o_{i,t}$ and the stepwise attention $A_{i,t}$ as follows: 
\begin{equation}
	\begin{split} 
		X_{i,t} = \text{Concat}(Q_{i,t}, A_{i,t}), \\
	\end{split}
\end{equation} 
where $X_{i,t}$ is the time-state representation of the trajectory to be predicted at time step $t$ in domain $i$. $\text{Concat}(\cdot)$ denotes the concatenation operation.

Pedestrian trajectories consist of time-sequence information, which is constructed from time-state information. Hence, the representation of all trajectories in an arbitrary domain $i$ can be constructed by time-state representation $X_{i,t}$ as follows:
\begin{equation}
	\begin{split} 
		F_i &= [\mathbf{f}_i^{(1)}, \mathbf{f}_i^{(2)}, ...,\mathbf{f}_i^{({N_i})}], \\
		\mathbf{f}_i^{(n)} &= [X_{i,1}^{n}, X_{i,2}^{n}, ..., X_{i,T_{obs}}^{n}], n\in [1, ..., {N_i}], \\
	\end{split}
\end{equation}
where $F_i$ is the feature space of all trajectories in domain $i$.
${N_i}$ is the number of pedestrians in domain $i$. 
$\mathbf{f}_i^{(n)}$ is the trajectory's pre-aligned representation of the $n^{th}$ pedestrian over $T_{obs}$ time steps in domain $i$.
$X_{i,t}^{n}$ is the time-state representation of the $n^{th}$ pedestrian's trajectory at time step $t$ in domain $i$.

\subsection{Recurrent Alignment}
As previously discussed, the goal of the recurrent alignment module is to eliminate the domain gap by extracting generalized trajectory features across different scenarios.
Hence, with the feature spaces $[F_1, F_2]$ of two source domains not aligned properly, we can align them by minimizing the discrepancy of the two source domains. 
Then, we can assume that the resulting features will be domain-invariant and generalized well on the unseen target domains.

\textbf{Recurrent Alignment Strategy.}
Different from traditional domain alignment strategies, the generalized pedestrian trajectory prediction task needs to consider the alignment at both the time-state and time-sequence levels.
Prior to alignment, we first use a multilayer gated recurrent unit (GRU) RNN to model the high-dimensional features of the trajectory at both time-state and time-sequence levels for prediction in domain $i$ based on hidden states.
Each RNN layer calculates the recurrent information in the input trajectory sequence as follows:
\begin{equation} \label{EQ5}
	\begin{split} 
		r_{i,t}^n &= \sigma(W_x^r X_{i,t}^n + W_h^r h_{i,t-1}^n +b^r), \\
		z_{i,t}^n &= \sigma(W_x^z X_{i,t}^n + W_h^z h_{i,t-1}^n +b^z), \\
		g_{i,t}^n &= \text{tanh}(W_x^g X_{i,t}^n + r_{i,t}^n \odot W_h^g h_{i,t-1}^n +b^g), \\
		h_{i,t}^n &= (1-z_{i,t}^n) \odot g_t^n + z_{i,t}^n \odot h_{i,t-1}^n, \\
	\end{split}
\end{equation}  
where $h_{i,t}^n$ is the hidden state of the trajectory's time-state representation $X_{i,t}^n$.
$r_{i,t}^n$, $z_{i,t}^n$, and $g_{i,t}^n$ are the reset, update, and new gates, respectively. 
$W$ and $b$ denote learning parameters. 
$\sigma$ is the activation function.
$\odot$ is the element-wise multiplication.
The initial hidden state $h_{i,0}$ is produced by a linear transformation of the feature embedding $\mathbf{F}_{i,t}^b$.
Hence, the feature space of each domain at different time steps can be represented as follows:
\begin{equation} \label{EQ6}
	\begin{split} 
		S_{i,t} =\sum_{n=1}^{N_i} h_{i,t}^{n},
	\end{split}
\end{equation} 
where $S_{i,t}$ is the feature space of all trajectories in domain $i$ at time step $t$.
Given $S_{i,t}$, we can minimize the discrepancy between the two source domains using the proposed recurrent alignment strategy and the recurrent alignment losses outlined in Algorithm~\ref{AL1}.
\textcolor{black}{$\mathcal{L}_{rec}(t)$ is the recurrent alignment loss $\mathcal{L}_{rec}$ at time step $t$.}
Then, our network is capable of modeling generalized trajectory features, which can be regarded as domain-invariant.	

\begin{algorithm}[t] 
	\SetAlgoLined 
	\caption{Recurrent Alignment Strategy}
	\label{AL1}
	\KwIn{Trajectory feature spaces of two source domains: 
		$F_1, F_2$, where:\quad \quad \quad \quad \quad 
		$F_i = [\mathbf{f}_i^{(1)}, \mathbf{f}_i^{(2)}, ...,\mathbf{f}_i^{({N_i})}]$,
		$\mathbf{f}_i^{(n)} = [X_{i,1}^{n},  ..., X_{i,T_{obs}}^{n}], n\in [1, ..., {N_i}]$, }
	
	$h_{i,t}^n \leftarrow X_{i,t}^n$   \algorithmiccomment{Equation~\ref{EQ5}} \\
	$S_{i,t} \leftarrow h_{i,t}^n$   \algorithmiccomment{Equation~\ref{EQ6}} \\
	\For{$t=1:T_{obs}$}{
		$\mathcal{L}_{rec}(t) \leftarrow [S_{1,t}, S_{2,t}]$  \algorithmiccomment{Equation~\ref{EQ7}} \\
	}
	$\mathcal{L}_{rec} \leftarrow \sum_{t=1}^{T_{obs}}\mathcal{L}_{rec}(t)$ \algorithmiccomment{Equation~\ref{EQ7}} \\
	Backward Loss $\mathcal{L}_{rec}$ \\
	Return 
\end{algorithm}

\textbf{Recurrent Alignment Loss.}
The alignment loss is to minimize the discrepancy between the source domains. 
To support the proposed recurrent alignment strategy, we design a recurrent alignment loss to minimize the discrepancy between the two source domains at both time-state and time-sequence levels as follows:
\begin{equation} 
	\begin{split} \label{EQ7}
		\mathcal{L}_{rec} &= \sum_{t=1}^{T_{obs}} ||S_{1,t}, S_{2,t}||_{\mathcal{H}}, \\
	\end{split}
\end{equation}
where $\mathcal{L}_{rec}$ is the recurrent alignment loss, which calculates the total discrepancy between the two source domains at each time step through the recurrent alignment strategy.
$||S_{1,t}, S_{2,t}||_{\mathcal{H}}$ is the alignment loss at time step $t$, indicating the discrepancy measurement between two source domains.
There are multiple choices for measuring such discrepancies, such as $L_2$ distance, MMD, and CORAL. 
We explore six measurement functions in Section~\ref{sec:eval} and choose $L_2$ distance loss for the generalized pedestrian trajectory prediction in this work.	

\begin{figure}[t]
	\centering
	\includegraphics[width=0.9\linewidth]{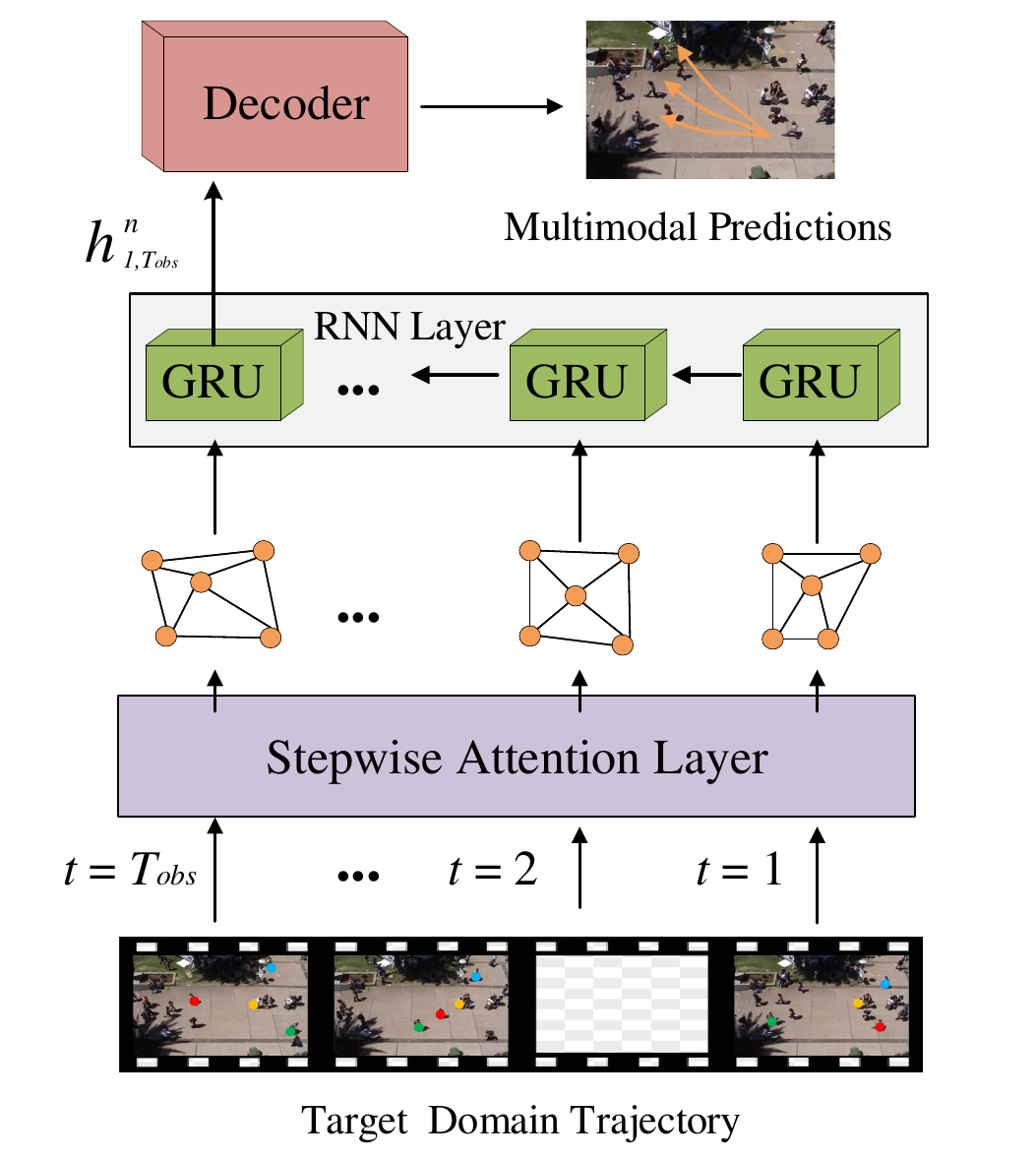}
	\caption{ 
		The inference phase of the proposed RAN framework.
		The input data are from target domain scenarios, and there is no requirement for data from at least two domains during the inference phase.
	}
	\label{fig4}
\end{figure}

\subsection{Prediction and Loss Function}
Pedestrian trajectory prediction aims to predict future trajectories that closely match the ground truth. Due to the multiple possibilities in the future, it is reasonable to have multiple potential future trajectories for a single observed trajectory. \textcolor{black}{
Following the previous best-of-K protocol, we use the mixture of expert (MoE) decoders to generate the corresponding $K$ possible predictive trajectories and choose the prediction that most closely matches the ground truth to calculate performance.}
Through the pre-aligned representation module and the recurrent alignment module, we can obtain the generalized trajectory features.
Then, we need to decode these features into predictions as follows:
\begin{equation} 
	\begin{split} \label{EQ8}
		(\mathbf{\hat{Y}}_i^n)^{K} = f_d(h_{i,T_{obs}}^n),
	\end{split}
\end{equation}
where $(\mathbf{\hat{Y}}_i^n)^{K}$ is $K$ predictions of the $n^{th}$ pedestrian in domain $i$ following previous best-of-$K$ metrics.
$f_d$ is the mixture of expert decoders implemented by $K$ MLP layers.
We use the $L_2$ distance loss as the prediction loss as follows:
\begin{equation} 
	\begin{split}
		\mathcal{L}_{pre} = \frac{1}{2 \times N_i}\sum_{i=1}^{2}\sum_{n=1}^{N_i} ||\mathbf{Y}_i^n - (\mathbf{\hat{Y}}_i^n)^{min}||_2, \\
	\end{split}
\end{equation}
where $\mathcal{L}_{pre}$ is the prediction loss of all trajectories in all source domains.
$(\mathbf{\hat{Y}}_i^n)^{min}$ denotes the prediction with the minimum error in $(\mathbf{\hat{Y}}_i^n)^{k}$.

The overall network is trained end-to-end by minimizing the total loss function $\mathcal{L}_{RAN}$ as follows:
\begin{equation} 
	\begin{split}
		\mathcal{L}_{RAN} = \lambda_1 \mathcal{L}_{rec} + \lambda_2 \mathcal{L}_{pre},
	\end{split}
\end{equation}
where $\lambda_1$ and $\lambda_2$ are coefficients of the loss function.

\subsection{Inference}
Because the two RNN layers share weights and the two stepwise attention layers share weights during training, the inference phase is straightforward, as illustrated in Figure~\ref{fig4}.
Given an observed trajectory from the target domain scenarios, we initially model its stepwise interaction at each time state using the stepwise attention layer. 
Subsequently, we combine this attention information with the trajectory embedding and input it into the well-aligned RNN layer to extract generalized trajectory features. 
Finally, we employ a multimodal trajectory decoder to decode the generalized trajectory features into the multimodal future predictions.

\textcolor{black}{
\subsection{Multiple Source Domains Discussion}
Theoretically, our method can handle an unlimited number of source domains in training by adapting the recurrent alignment loss. Specifically, when the number of source domains exceeds two, the recurrent alignment loss $\mathcal{L}_{recm}$ is calculated as follows:
\begin{equation} 
	\begin{split} \label{EQ7}
		\mathcal{L}_{recm} &= \frac{1}{m^2}\sum_{t=1}^{T_{obs}}\sum_{p,q=1}^{m} ||S_{p,t}, S_{q,t}||_{\mathcal{H}}, \\
	\end{split}
\end{equation}
where $||S_{p,t}, S_{q,t}||_{\mathcal{H}}$ is the alignment loss between two source domains $p$ and $q$ at time step $t$. $m$ is the number of source domains.
However, as the number of source domains increases, the computation complexity of the recurrent alignment loss also increases. 
Specifically, during training, if there are $m$ source domains, and each source domain has a trajectory observation time of $T_{obs}$, then the computational complexity of the recurrent alignment loss $\mathcal{L}_{recm}$ is $O(T_{obs} \cdot m^2)$.
}

\section{Experiments and Discussions} \label{sec:eval}
	\begin{table*}[t]
	\caption{Comparison for generalization results (ADE/FDE) on ETH-UCY. The model is trained in SDD and NBA. $\ast$ means the results are reproduced using the officially released code. \textbf{Bold} indicates the best performance. The lower the better.  }
	\centering
	\resizebox{1\linewidth}{!}{
		\setlength{\tabcolsep}{1em}%
		\begin{tabular}{c|cccccc|c}
			\toprule
			\makecell{Target\\ Domain}   & \makecell{Social-STG \\ CNN$^\ast$~\cite{mohamed2020social}}  &\makecell{SGCN$^\ast$ \\ \cite{shi2021sgcn}}& \makecell{TPNMS$^\ast$  \\ \cite{liang2021temporal}}  & \makecell{GP-\\Graph$^\ast$~\cite{bae2022learning} }   &  \makecell{Social-\\VAE$^\ast$~\cite{xu2022socialvae}} & \makecell{Graph-\\TERN$^\ast$~\cite{bae2023set} }     & \multirow{1}*{\textbf{Ours}}      \\
			\midrule
			ETH		  &  0.66/0.92      & 0.72/1.16    & 0.69/0.97   & 0.54/0.76 	  & 0.57/0.79 &   0.53/0.74 &     \textbf{0.41}/\textbf{0.69} \\
			
			UCY	      & 0.68/0.99     & 0.52/0.79    & 0.63/0.90   & 0.27/0.63 	  & 0.45/0.66 &   0.25/0.58 &     \textbf{0.13}/\textbf{0.21} \\
			UNIV      & 0.61/0.89     & 0.51/0.77    & 0.60/0.87   & 0.37/0.71 	  & 0.53/0.75 &   0.37/0.69 &     \textbf{0.25}/\textbf{0.46} \\
			
			ZARA01    & 0.63/1.02     & 0.58/0.90    & 0.61/0.99   &0.36/0.59 	  & 0.50/0.76 &   0.34/0.57 &     \textbf{0.22}/\textbf{0.41} \\
			
			ZARA02	  & 0.61/1.01     & 0.54/0.78    & 0.60/0.97   & 0.29/0.42 	  & 0.44/0.65 &   0.29/0.43 &     \textbf{0.16}/\textbf{0.31} \\
			\midrule
			Average   & 0.64/0.97     & 0.57/0.88    & 0.63/0.94   & 0.37/0.62 	  & 0.49/0.72 &   0.36/0.60 &     \textbf{0.23}/\textbf{0.42} \\
			
			%
			%
			%
			\bottomrule
		\end{tabular}
	}
	
	\label{table1}
\end{table*}

\begin{table*}[t]
	\caption{Comparison for generalization results (ADE/FDE) on SDD. The model is trained on ETH-UCY and NBA. $\ast$ means the results are reproduced using the official released code. \textbf{Bold} indicates the best performance. The Lower the better.  }
	\centering
	\resizebox{1\linewidth}{!}{
		\setlength{\tabcolsep}{1em}%
		\begin{tabular}{c|cccccc|c}
			\toprule
			\makecell{Target\\ Domain}   & \makecell{Social-STG \\ CNN$^\ast$~\cite{mohamed2020social}}  &\makecell{SGCN$^\ast$ \\ \cite{shi2021sgcn}}& \makecell{TPNMS$^\ast$  \\ \cite{liang2021temporal}}  & \makecell{GP-\\Graph$^\ast$~\cite{bae2022learning} }   &  \makecell{Social-\\VAE$^\ast$~\cite{xu2022socialvae}} & \makecell{Graph-\\TERN$^\ast$~\cite{bae2023set} }     & \textbf{Ours}      \\
			\midrule
			SDD 	  & 23.24/36.49     & 20.44/32.53    & 25.43/38.56   & 16.79/27.02 	  & 17.12/28.06 &   14.76/25.47 &     \textbf{10.97}/\textbf{19.95} \\
			\bottomrule
		\end{tabular}
	}
	\label{table2}
\end{table*}

\begin{table}[t]
	\caption{Dataset split for train/test/whole sets for three different datasets. Trajectories in the train set and test set do not overlap.}
	\centering
	\resizebox{1\linewidth}{!}{
		\setlength{\tabcolsep}{1em}%
		\begin{tabular}{c|ccc}
			\toprule
			\multirow{2}*{Dataset} & \multicolumn{3}{c}{Number of Trajectories}\\
			\cmidrule(r){2-4}\
			& Training Set & Test Set & Whole Set\\
			\midrule
			ETH-UCY 		&	34,161		&	3,109		&	37,270		 	\\
			SDD 		&	8,494		&	2,829		&	11,323 		\\
			NBA 		&	205,970	&	51,260	& 	257,230		\\
			
			\bottomrule
		\end{tabular}
	}
	\label{table_split}
\end{table}

In this section, we evaluate the pedestrian trajectory prediction performance of our proposed RAN and compare it with existing state-of-the-art methods on four standard benchmark datasets. Additionally, we conduct detailed ablation studies to explore the contribution of each component.
\subsection{Evaluation Datasets.}
We evaluate our proposed method on three widely used benchmark datasets, each situated in different scenarios and representing different domains.

\textbf{ETH-UCY.}
The ETH-UCY dataset~~\cite{lerner2007crowds, pellegrini2009you} is captured in the scenes of outdoor pedestrian paths. 
It contains trajectories of 1,536 pedestrians in four different scenes, \textit{i.e.}, ETH, HOTEL, UNIV, ZARA. 
It contains five sub-datasets captured from downfacing surveillance cameras annotated at 2.5Hz.
To ensure a fair evaluation, we follow the previous leave-one-out protocol widely adopted in previous studies~\cite{xu2022socialvae, bae2023set}, where it is trained on four subsets and tested on the remaining one.
The trajectories within the ETH-UCY dataset are documented in world coordinates, and thus we report the results in meters as previous methods~\cite{xu2022socialvae, bae2023set}. 

\textbf{SDD.}
The Stanford Drone Datase~(SDD)~~\cite{robicquet2016learning} is captured in the scenes of the university campus. 
It is a well-established benchmark for human trajectory prediction in bird’s eye view, which consists of 20 subsets captured using a drone.
It collects multi-agent trajectories of various types, including pedestrians, bicyclists, skateboarders, cars, buses, and golf carts, in a university campus setting. The dataset is notable for its vast size, containing over 11,000 individual pedestrian trajectories, which result in more than 185,000 interactions between pedestrians and over 40,000 interactions between pedestrians and their surroundings. To ensure a fair and consistent evaluation of our method, we adhere to the standard training and testing splits used in previous studies~\cite{xu2022socialvae}.
Trajectories in SDD are documented in camera coordinates, and hence, we report the results in pixels as previous approaches~\cite{xu2022socialvae, bae2023set}. 

\textbf{NBA.}
The SportVU NBA movement~(NBA) dataset~~\cite{yue2014learning} is captured in the scenes of basketball courts. It is a more challenging benchmark that contains rich social interactions.
Due to the large size of the original dataset, we randomly extracted 257,230 trajectories in the “Rebound” subset following the previous method~\cite{xu2022socialvae}.
The extracted trajectories capture a rich set of agent-agent interactions and highly non-linear motions. It is important to note that the overall frequency and the adversarial and cooperative nature of the interactions in this dataset are significantly different from those in ETH/UCY and SDD~\cite{xu2022socialvae}.
Trajectories in the NBA dataset are documented in world coordinates, and hence, we report the results in meters as the previous approach~\cite{xu2022socialvae}. 

\begin{table*}[t]
	\caption{Comparison for generalization results (ADE/FDE) on NBA. The model is trained on ETH-UCY and SDD. $\ast$ means the results are reproduced using the official released code. \textbf{Bold} indicates the best performance. The Lower the better.  }
	\centering
	\resizebox{1\linewidth}{!}{
		\setlength{\tabcolsep}{1em}%
		\begin{tabular}{c|cccccc|c}
			\toprule
			\makecell{Target\\ Domain}   & \makecell{Social-STG \\ CNN$^\ast$~\cite{mohamed2020social}}  &\makecell{SGCN$^\ast$ \\ \cite{shi2021sgcn}}& \makecell{TPNMS$^\ast$  \\ \cite{liang2021temporal}}  & \makecell{GP-\\Graph$^\ast$~\cite{bae2022learning} }   &  \makecell{Social-\\VAE$^\ast$~\cite{xu2022socialvae}} & \makecell{Graph-\\TERN$^\ast$~\cite{bae2023set} }     & \textbf{Ours}      \\
			\midrule
			NBA	      & 3.16/6.07     & 3.28/6.97    & 4.32/8.76   & 4.57/8.23 	  & 2.11/4.54 &   2.32/4.83 &     \textbf{1.28}/\textbf{2.64} \\
			\bottomrule
		\end{tabular}
	}
	\label{table3}
\end{table*}

\begin{table*}[t]
	\caption{Ablation study of each component of our method in ADE/FDE in meters on ETH-UCY. The model is trained in SDD and NBA. The lower the better.}
	\centering
	\resizebox{1\linewidth}{!}{
		\setlength{\tabcolsep}{1em}%
		\begin{tabular}{ccc|cccccc}
			\toprule
			Variants & PAR & RA \quad & ETH & HOTEL & UNIV & ZARA01 & ZARA02 & Average  \\
			\midrule
			(1)	 		&  	\ding{55}  &  \ding{55}	 & 0.50/0.81 & 0.21/0.40 & 0.31/0.58 & 0.29/0.57 & 0.20/0.38 & 0.30/0.55	 \\
			(2)			&  	\ding{55}  &   \checkmark   & 0.43/0.73 & 0.16/0.29 & 0.30/0.57 & 0.27/0.54 & 0.19/0.37 & 0.27/0.50	 \\
			(3) 		&   \checkmark  &   \ding{55}   & 0.46/0.77 & 0.18/0.33 & 0.26/0.47 & 0.23/0.45 & 0.18/0.32 & 0.26/0.47		 \\
			(4)  &  \checkmark   &   \checkmark     & \textbf{0.41}/\textbf{0.69} & \textbf{0.13}/\textbf{0.21} & \textbf{0.25}/\textbf{0.46} & \textbf{0.22}/\textbf{0.41}	 & \textbf{0.16}/\textbf{0.31} & \textbf{0.23}/\textbf{0.42}	 \\
			\bottomrule
		\end{tabular}
	}
	
	\label{table4}
\end{table*}

\subsection{Experimental Setting}
\textbf{Evaluation Protocol.}
We treat the three datasets, \textit{i.e.}, ETH-UCY, SDD and NBA datasets, in different scenarios as different domains.
For generalized pedestrian trajectory prediction, the model can not reach the target domain data in the training phase.
Specifically, taking source domains A and B and target domains C as examples, we split A, B and C into train sets and test sets, respectively. All train sets and the test set do not overlap.
We train and validate the model on train and test sets of A and B, respectively. Then, we test the model on the test set of C.
More details can be found in Tabble~\ref{table_split}.

\textbf{Evaluation Metrics.}
Following previous methods~\cite{xu2022socialvae, xu2022adaptive}, we employ two commonly used metrics, \textit{i.e.}, Average Displacement Error (ADE) and Final Displacement Error (FDE), as our evaluation metrics.
ADE means the average $L_2$ distance between all points of the prediction and ground truth. 
FDE means the $L_2$ distance between the destination points of the prediction and ground truth.

\textbf{Implementation Details.}
Similar with the previous methods~\cite{xu2022adaptive,xu2022socialvae,mohamed2020social}, we calculate the best-of-20 performance.
We observe 8 frames for each trajectory and predict the subsequent 12 frames.
The two MLP encoders are all 2-layer linear transformations with the sigmoid activation function, and the hidden sizes are set to 64 and 128.
The dimensions of the learnable parameters $W_{i,t}^{q}$, $W_{i,t}^{k}$, and $W_{i,t}^{v}$ in attention are all set to 256.
We employ GRUs as the RNN structure.
The activation function $\sigma$ is sigmoid.
The coefficients $\lambda_1$ and $\lambda_2$ in the loss function are set to 1.
The Adam optimizer is used to train our model by 300 epochs with a learning rate of 0.001 and batch size of 512, decaying by 0.5 with an interval of 50, on one GTX-3090 GPU.

\begin{table*}[t]
	\caption{\textcolor{black}{Comparison for performance (ADE/FDE) on ETH-UCY. Results are trained on different source domains and tested on the target domain ETH-UCY. \textbf{Bold} indicates the best performance. The lower the better. }}
	\centering
	\resizebox{1\linewidth}{!}{
		\setlength{\tabcolsep}{0.6em}%
		\begin{tabular}{c|c|cccccc|c}
			\toprule
			&& \makecell{Social-STG \\ CNN$^\ast$~\cite{mohamed2020social}}  &\makecell{SGCN$^\ast$ \\ \cite{shi2021sgcn}}& \makecell{TPNMS$^\ast$  \\ \cite{liang2021temporal}}  & \makecell{GP-\\Graph$^\ast$~\cite{bae2022learning} }   &  \makecell{Social-\\VAE$^\ast$~\cite{xu2022socialvae}} & \makecell{Graph-\\TERN$^\ast$~\cite{bae2023set} }     & \textbf{Ours}      \\
			\hline			
			Case 1&\makecell{Source: ETH-UCY \\ Target: ETH-UCY}   &   0.44/0.75   &   0.37/0.65  &  0.38/0.73  &  	0.23/0.39  & 0.24/0.42 &  0.24/0.38  &  \textbf{0.21/0.36}    \\
			\hline
			Case 2&\makecell{Source: ETH-UCY \& SDD \\ Target: ETH-UCY}   &  0.44/0.75   &  0.35/0.65  &  0.38/0.73  &  0.23/0.39	  & 0.23/0.41 &   0.23/0.38 &   \textbf{0.20/0.36}   \\
			\hline
			Case 3&\makecell{Source: NBA \& SDD \\ Target: ETH-UCY}   & 0.64/0.97     & 0.57/0.88    & 0.63/0.94   & 0.37/0.62 	  & 0.49/0.72 &   0.36/0.60 &     \textbf{0.23}/\textbf{0.42} \\	
			\bottomrule
		\end{tabular}
	}
	\label{case}
\end{table*}
\begin{table*}[th]
	\caption{Ablation study of recurrent alignment loss on ETH-UCY dataset. The model is trained in SDD and NBA. The lower the better.}
	\centering
	\resizebox{1\linewidth}{!}{
		\setlength{\tabcolsep}{2.2em}%
		\begin{tabular}{c|ccccc|c}
			\toprule
			& MMD & CORAL & KLD & JS & COS & L2~(Ours)  \\
			\midrule
			ADE   & 0.32 & 0.30 & 0.25 & 0.27 & 0.32 & \textbf{0.23}	 \\
			FDE   & 0.58 & 0.55 & 0.45 & 0.49 & 0.56 & \textbf{0.42}	 \\
			\bottomrule
		\end{tabular}
	}	
	\label{table5}
\end{table*}

\subsection{Quantitative Analysis}
Tables~\ref{table1}-\ref{table3} show the performance compared to six state-of-the-art methods in three widely used datasets, which are regarded as three different domains.
The model is trained in two selected domains and tested on the remaining one to evaluate cross-domain generalization. 
Consistent with established protocols, we express the ETH-UCY and NBA results in meters and the SDD results in pixels.
To eliminate the impact of different units of measurement on the model, we unified them as meters during training.
\textcolor{black}{Given our emphasis on generalization ability and a distinct experimental setup compared to previous state-of-the-art methods, we reproduced previous state-of-the-art methods under the generalization setting through their officially released codes without post-processing to ensure fair comparisons.}

\textbf{ETH-UCY.}
Table~\ref{table1} shows the generalization performance on the ETH-UCY dataset, which is trained on the SDD and NBA source domains.
The experimental results show that our method achieves the best performance in both the ADE and FDE metrics.
Specifically, our method improves the performance of the previous best method, Graph-TERN~\cite{bae2023set}, from $0.36$ to $0.23$ on average ADE and from $0.60$ to $0.42$ on average FDE.  
Besides, we also outperform other approaches in all five subsets on both ADE and FDE metrics, which demonstrates our superior generalization capability.
The underlying reason is that our recurrent alignment strategy enables the model to learn domain-invariant features across different scenarios.

\textbf{SDD.}
Table~\ref{table2} shows the generalization performance on the SDD dataset, which is trained on the ETH-UCY and NBA source domains.
The experimental results show that our method achieves the best performance in both the ADE and FDE metrics compared to the other methods.
Specifically, our method improves performance over the previous best method, Graph-TERN~\cite{bae2023set}, from $14.76$ to $10.97$ in the average ADE metric and from $25.47$ to $19.95$ in average FDE metric. This demonstrates that our method has superior generalization capability compared to other approaches.

\textbf{NBA.}
Table~\ref{table3} shows the generalization performance on the NBA dataset, which is trained on ETH-UCY and SDD source domains.
The experimental results show that our method achieves the best performance in both ADE and FDE metrics.
Specifically, our method improves the average ADE/FDE from $1.28/2.64$ to $2.11/4.45$ compared to the previous best performance method, SocialVAE~\cite{xu2022socialvae}.
Notably, NBA dataset is captured in the basketball court, which presents denser interactions and a larger domain gap compared to the other datasets. 
Our method achieves more performance improvement facing such a larger domain gap, \textit{i.e.} $44.8\%/45.3\%$~(NBA), $36.1\%/30\%$~(ETH-UCY) and $25.6\%/21.6\%$ (SDD), compared to Graph-TERN~\cite{bae2023set}.
These results underscore our method's exceptional ability to generalize across varied and complex domains for pedestrian trajectory prediction.

\begin{figure}[t]
	\centering
	\includegraphics[width=1\linewidth]{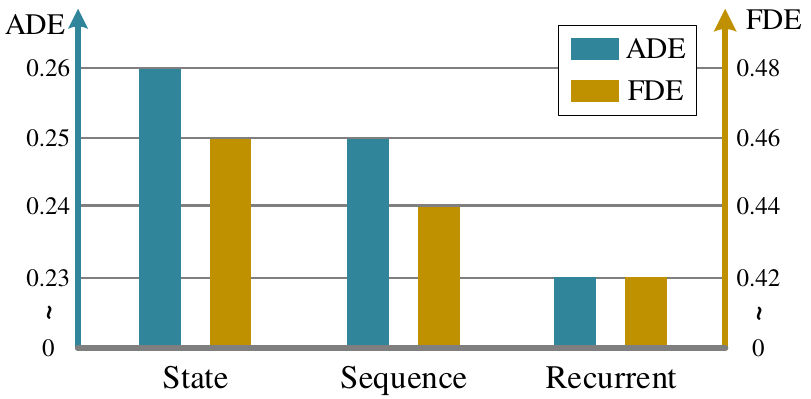}
	\caption{Ablation study of different alignment approaches on ETH-UCY dataset. The model is trained on SDD and NBA. The lower the better.}
	\label{abl_fig1}
\end{figure}

\subsection{Ablation Study}
In this section, we first study the contribution of the important modules of RAN.
Then we evaluate the impact of different alignment approaches.
Subsequently, we study the impact of different alignment units and coefficients of our loss function.
Finally, we evaluate the efficiency of our method.

\textbf{Contribution of Each Component.} 
We conduct experiments to evaluate the major components of our RAN on the ETH-UCY dataset, as shown in Table~\ref{table4}.
The recurrent alignment module and the pre-aligned representation module are the two major contributions of our proposed framework. 
PAR and RA denote the pre-aligned representation module and the recurrent alignment module, respectively. 
To isolate the effect of each component and determine its significance to the overall performance of the model, removing PAR denotes that we only embed the target trajectory for recurrent alignment at each timestamp.
Furthermore, we remove RA to evaluate our recurrent aligned strategy.

\begin{figure}[t]
	\centering
	\includegraphics[width=1\linewidth]{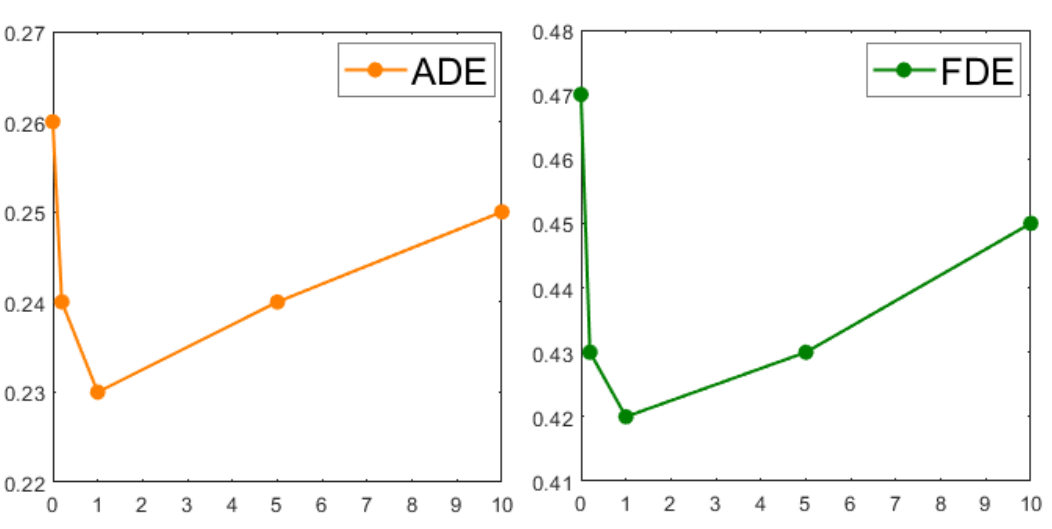}
	\caption{Ablation study of using different loss coefficients on ETH-UCY dataset. The model is trained on SDD and NBA.
		X-axis is $\lambda_1 / \lambda_2$. Y-axis is ADE or FDE. The lower the better.}
	\label{abl_fig2}
\end{figure}

The experimental results show that removing each component results in performance degradation. Specifically, removing PAR results in a performance decrease from 0.23/0.42 to 0.26/0.47 on average ADE/FDE metrics. Removing RA results in a performance decrease from 0.23/0.42 to 0.27/0.50 on average ADE/FDE metrics. Removing both PAR and RA results in a performance decrease from 0.23/0.42 to 0.30/0.55 on average ADE/FDE. These results underscore that both the recurrent alignment module and pre-aligned representation module are effective in improving the generalization capability of our method.

\begin{figure*}[t]
	\centering
	\includegraphics[width=1\linewidth]{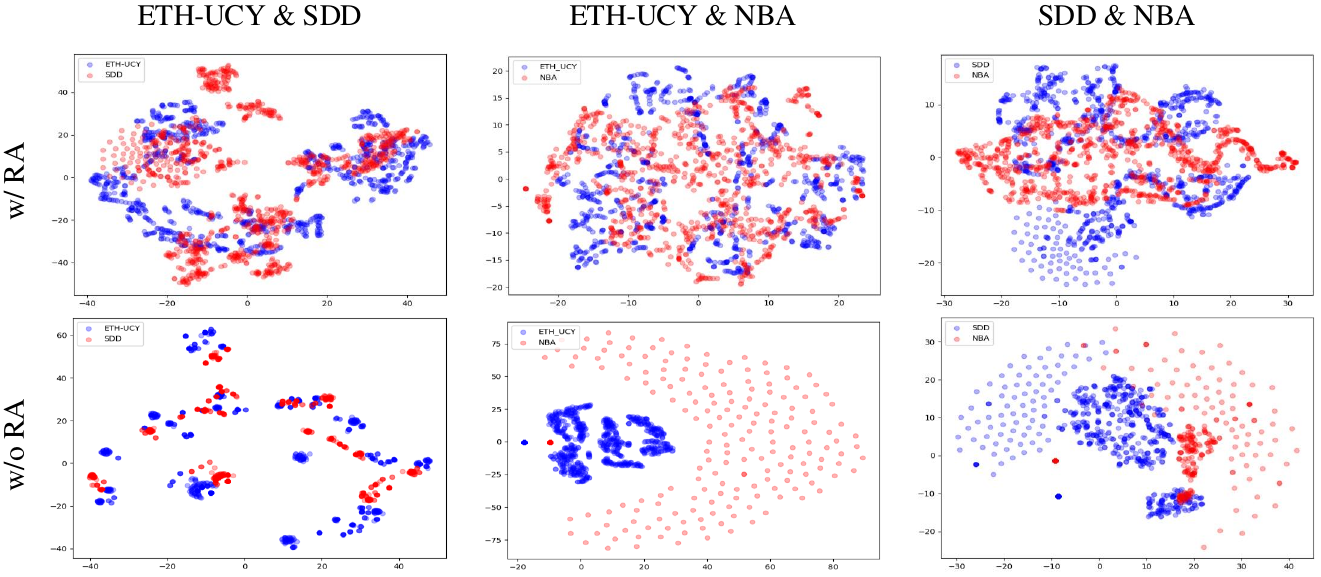}
	\caption{ 
		Visualization of feature spaces of different domains using t-SNE.
		“A \& B” on the top row denotes that the source domains are A and B during training.
		The blue and red dots denote feature spaces of the corresponding source domains A and B, respectively.
		To differentiate point overlap, we set the transparency value of points to 0.3. 
		Deep colour regions indicate points overlap.
	}
	\label{visualization}
\end{figure*}

\begin{figure*}[t]
	\centering
	\includegraphics[width=1\linewidth]{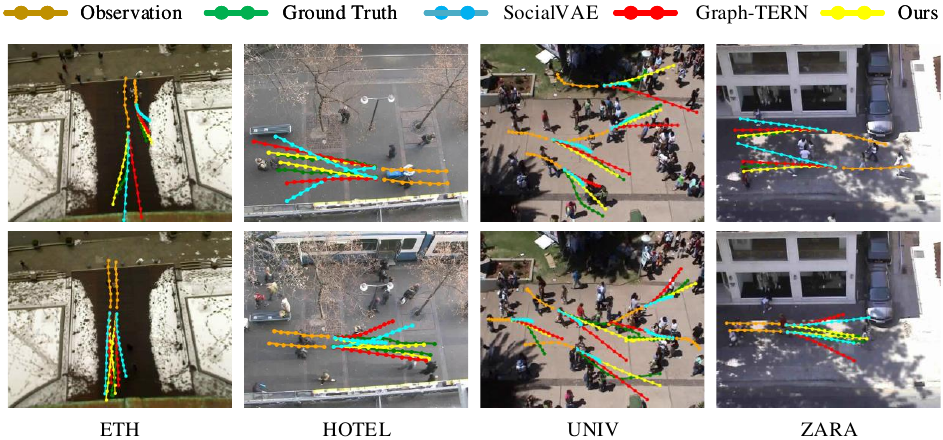}
	\caption{ 
		Visualization of predicted trajectories compared with SocialVAE~\cite{xu2022socialvae} and Graph-TERN~\cite{bae2023set} on ETH-UCY dataset.  The model is trained on SDD and NBA. Given the observed trajectories, we illustrate the ground truth paths and predicted trajectories by ours, SocialVAE and Graph-TERN in four different scenes. We see that our results are much closer to the ground truth compared with the other two approaches.
	}
	\label{vis_trj}
\end{figure*}

\textcolor{black}{
\textbf{Analysis of Different Source Domain.} }
\textcolor{black}{As shown in Table~\ref{case}, we conduct experiments to use different source domains in training to validate the effectiveness of our method.
The experimental results demonstrate that our method achieves the best performance under all three settings. }

\textcolor{black}{The reason for our superior performance in “Case 1” is that, even within the same trajectory dataset, “out-of-distribution” data can exist. These out-of-distribution data are caused by the long-tail distribution phenomenon in trajectory data. Our approach can mitigate the domain shift among out-of-distribution data, leading to improved performance.}

\textcolor{black}{The reason for our superior performance in “Case 3” is that we address the domain shift problem by employing a recurrent aligned method specifically tailored to trajectory data. This approach aligns the feature spaces of different domains, enabling the encoder to extract common features, which enhances the model’s generalization capability.}

\textcolor{black}{Comparing “Case 1” with “Case 2,” the experimental results show that a small expansion of the training data (from ETH-UCY to ETH-UCY \& SDD) leads to a slight performance improvement. This is because both “Case 1” and “Case 2” include the ETH-UCY dataset in their training sets, allowing the model to learn from data in the same domain as the test set. “Case 2” further expands this by adding additional data, serving as data expansion, which accounts for the slight performance enhancement.}

\textcolor{black}{Comparing “Case 1” with “Case 3,” the experimental results show that “Case 1” performs better. This is because, although “Case 3” has a larger amount of data, its training set does not include data from the same domain as the ETH-UCY test set, meaning it has not encountered data from the same domain during training. In contrast, both the training and test sets in “Case 1” come from the same domain. Thus, “Case 1” shows better performance than “Case 3”.}

\textcolor{black}{Additionally, the performance drop of our method before and after the domain shift is the smallest compared to other methods, further demonstrating the effectiveness of our approach in addressing the domain generalization problem.}

\begin{table}[t]
	\caption{Ablation study about numbers of parameters and inference time for 100 trajectories on NBA dataset.}
	\centering
	\resizebox{1\linewidth}{!}{
		\setlength{\tabcolsep}{1em}%
		\begin{tabular}{c|cc}
			\toprule
			& Inference time	&  \# of Parameters\\
			\midrule
			
			SocialVAE~\cite{xu2022socialvae} 		&	13.6ms		&	2.144M			 	\\	
			\hline
			RAN (Ours)		&	1.9ms		&	2.222M			 	\\	
			\bottomrule
		\end{tabular}
	}
	\label{tab_effi}
\end{table}

\textbf{Analysis of Different Alignment Approaches.}
As shown in Figure~\ref{abl_fig1}, we replace our recurrent alignment strategy in RAN with time-state and time-sequence alignment on the ETH-UCY dataset. 
“State” denotes the time-state alignment strategy, implemented by aligning the feature space in each step of the attention layers.
“Sequence” denotes the time-sequence alignment strategy, implemented by only aligning the feature space at the last step in the RNN layers.
“Recurrent” denotes our recurrent alignment strategy, which considers both time-state and time-sequence alignment.

Experiment results show that our recurrent alignment strategy achieves the best performance compared to the time-state and time-sequence alignment strategies.
Specifically, compared to the time-state alignment strategy, we improve the average ADE/FDE from 0.26/0.46 to 0.23/0.42. 
Compared with the time-sequence alignment strategy, we improve the average ADE/FDE from 0.25/0.44 to 0.23/0.42. 
This indicates the superiority of our method in domain alignment for generalized pedestrian trajectory prediction.
Furthermore, the performance of the time-sequence strategy exceeds that of the time-state alignment strategy, demonstrating the importance of extracting temporal information for pedestrian trajectory prediction tasks.

\textbf{Analysis of Recurrent Alignment Loss.} 
As shown in Table~\ref{table5}, we conduct experiments to evaluate the impact of using different alignment losses as the alignment unit in $\mathcal{L}_{rec}$ on ADE/FDE metrics on ETH-UCY dataset.
Specifically, MMD, CORAL, KLD, JS, and COS denote using MMD loss, CORAL loss, Kullback-Leibler divergence loss, Jensen–Shannon divergence loss, and cosine similarity as the alignment unit in $\mathcal{L}_{rec}$, respectively.

The evaluation results show that the performance of using the $L_2$ loss exceeds that of using other loss functions. It suggests that the $L_2$ distance is a more suitable domain distance measurement metric for pedestrian trajectory prediction tasks. 
As mentioned in Section~\ref{sec:method}, we use the $L_2$ distance as the alignment unit in $\mathcal{L}_{rec}$ in this work.
One possible explanation is that high-dimensional feature representations may still preserve spatial-level information, \textit{i.e.}, the spatial distance, in pedestrian trajectory prediction tasks.

\textbf{Analysis of Loss Coefficient.}
The overall loss function of our RAN consists of two components, recurrent alignment loss $\mathcal{L}_{rec}$ and prediction loss $\mathcal{L}_{pre}$. Since improving generalization can lead to decreased accuracy in the source domain, the weight coefficients for $\mathcal{L}_{rec}$ and $\mathcal{L}_{pre}$ represent the balance between generalization and accuracy in the model, which is important for model training.

As shown in Figure~\ref{abl_fig2}, we evaluate different loss coefficients to validate the appropriate proportion of the alignment module in our framework on the ETH-UCY dataset.
The proportion of the alignment module is represented by the figure's loss coefficient ratio $C = \lambda_1 / \lambda_2$.
The evaluation results indicate that an excessive proportion of alignment modules sacrifices the model's prediction accuracy, resulting in a performance decline. 
Conversely, a deficient proportion of alignment modules diminishes the model's generalization capacity, leading to a performance decrease.
Based on the experimental results, we choose $C = 1$ in this work, which achieves the best performance and represents a balance between generalization and precision for the pedestrian trajectory prediction task.
\textcolor{black}{In practical applications, if the target scenario is highly similar to the source domain, we choose a smaller $\lambda_1$ and a larger $\lambda_2$. Conversely, if the target scenario differs significantly from the source domain, we select a larger $\lambda_1$ and a smaller $\lambda_2$.}

\textbf{Analysis of Efficiency.}
We conduct experiments to compare the inference time and the number of parameters with SocialVAE~\cite{xu2022socialvae} on NBA dataset. Since the inference time for each trajectory is very short, our experimental results provide the inference time for 100 pedestrian trajectories.

The experimental results show that while our model has a comparable number of parameters to SocialVAE, our inference time is significantly shorter. The reason for the smaller number of parameters in our model is that, although we require two sets of stepwise attention layers and RNN layers for the two different source domains during training, they share the same parameters.
There are two reasons for our faster inference time. First, during the inference phase, we only need a single set of stepwise attention layers and RNN layers for the target domain. Second, our network is more lightweight. Specifically, we use the mixture of expert decoders composed of multiple MLPs to generate multimodal trajectories, rather than using the variational auto-encoder as in previous methods.
These experiment results demonstrate that our method is not only effective but also simple and efficient.

\section{Qualitative Analysis}
\textbf{Feature Space Visualization.}
As illustrated in Figure~\ref{visualization}, we visualize feature spaces in different domains using t-SNE.
“w/ RA” and “w/o RA” denote with and without the RA module, respectively.
In each figure, we randomly select a test batch data for each domain.
Each point in the figure represents the feature of one trajectory.
Points in different colours denote features in different domains.
Among these three sets of figures, there is a notably higher feature overlap when using the RA module, demonstrating that our alignment strategy can learn generalized trajectory features across different domains. 
Notably, due to dense interaction, NBA has a more extensive domain gap with ETH-UCY and SDD.
The results show that our approach makes feature alignment more obvious in the face of more significant domain gaps, such as ETH-UCY \& NBA and SDD \& NBA.
This also demonstrates the significant generalization capability of our method.

\textbf{Predicted Trajectory Visualization.}
As shown in Figure~\ref{vis_trj}, we compare the visualized trajectories in four different scenes with the best performance methods, SocialVAE~\cite{xu2022socialvae} and Graph-TERN~\cite{bae2023set}, on ETH-UCY dataset.
The model is trained in the source domains SDD and NBA.
The results in four different scenes show that our method surpasses the trajectory prediction accuracy of both SocialVAE and Graph-TERN. 
After transferring from the source domain to the target domain, the predictive performance of both SocialVAE and Graph-TERN decreases significantly. Although they can still predict the general trend of future movements, their accuracy is insufficient.
In contrast, our method can still accurately predict future trajectories even after domain changes.
This enhanced accuracy is particularly evident in the finer details of the predicted trajectories.
This enhancement in prediction accuracy after domain change can be attributed to the effective feature alignment employed in our approach.

\section{Conclusion} \label{sec:conc}

In this paper, we propose and study a new task named generalized pedestrian trajectory prediction. 
This task aims to generalize the model to unseen domains without accessing the target domain trajectories in the training process.
Furthermore, we propose the Recurrent Aligned Network~(RAN), a novel alignment framework tailed for generalized pedestrian trajectory prediction.
In RAN, we devise a recurrent alignment module with a recurrent alignment loss to effectively align different trajectory domains at both the time-state and time-sequence levels. 
In addition, we devise a pre-aligned representation module to consider human interactions during the domain alignment process.
The experimental results unequivocally highlight the superior generalization capabilities of our method.
To the best of our knowledge, our study is the first to explore the generalized setting for pedestrian trajectory prediction, offering significant insights for real-world autonomous driving systems. 

\textbf{Limitation and Future Work.} 
One major limitation of our method is that we need two source domains in the training phase due to the characteristics of the domain alignment approach.
In the future, we plan to tackle this problem by investigating a single-domain generalization solution for generalized pedestrian trajectory prediction.



\ifCLASSOPTIONcaptionsoff
  \newpage
\fi

\bibliographystyle{IEEEtran}
\bibliography{IEEEabrv,referemce}

\begin{IEEEbiography}[{\includegraphics[width=1in,height=1.25in,clip,keepaspectratio]{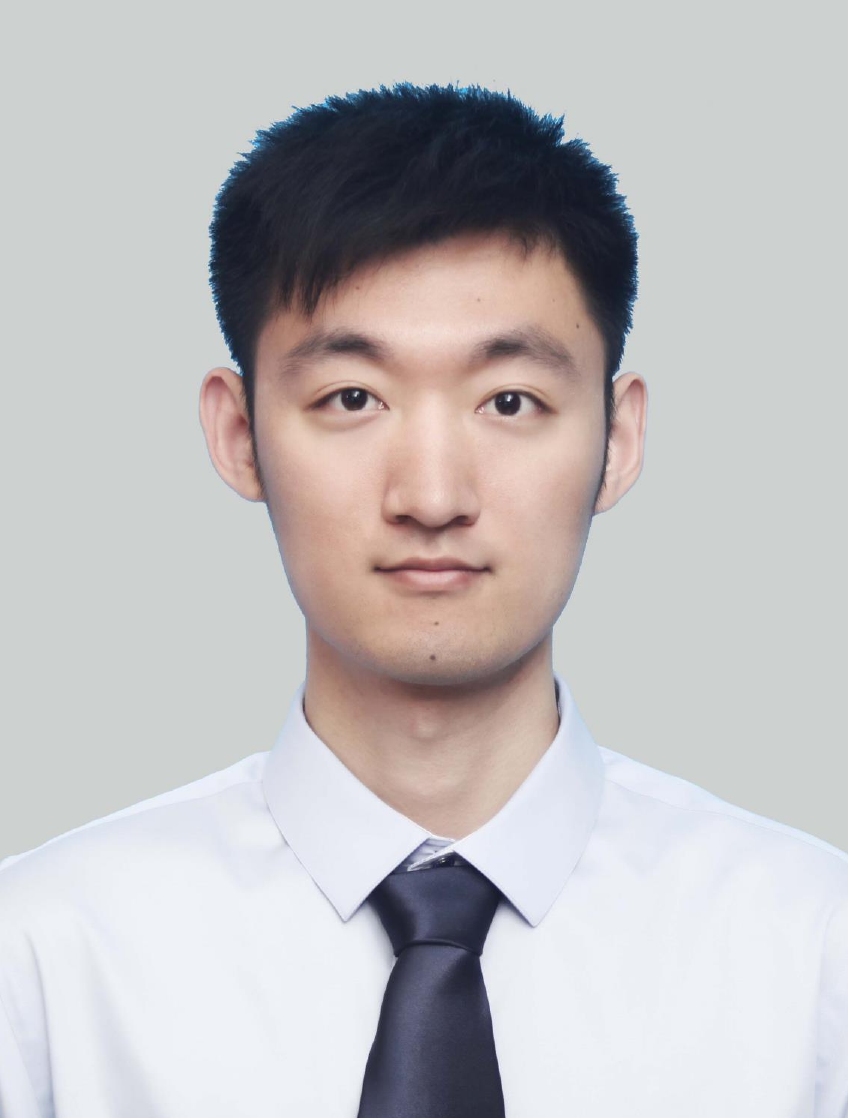}}]{Yonghao Dong}
	received the B.S. degree in Measurement Control Technology and Instrument from East China University of Science and Technology, Shanghai, China, in 2018, and the M.S. degree in Electrical and Computer Engineering from the University of Illinois at Chicago, Chicago, Illinois, USA, in 2019.
	He is currently pursuing the Ph.D. degree at the Institute of Artificial Intelligence and Robotics, Xi'an Jiaotong University, Xi'an, China. His research interests include computer vision and autonomous driving.	
\end{IEEEbiography}

\begin{IEEEbiography}[{\includegraphics[width=1in,height=1.25in,clip,keepaspectratio]{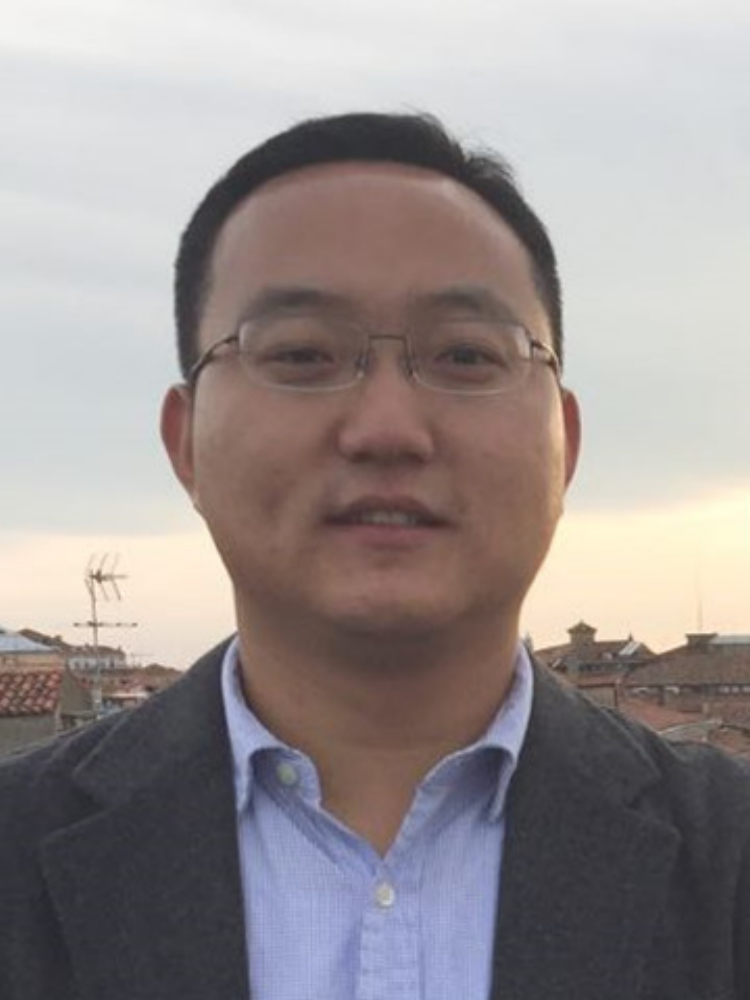}}]{Le Wang}(Senior Member, IEEE) received the B.S. and Ph.D. degrees in Control Science and Engineering from Xi'an Jiaotong University, Xi'an, China, in 2008 and 2014, respectively. From 2013 to 2014, he was a visiting Ph.D. student with Stevens Institute of Technology, Hoboken, New Jersey, USA. From 2016 to 2017, he was a visiting scholar with Northwestern University, Evanston, Illinois, USA. He is currently a Professor with the Institute of Artificial Intelligence and Robotics, Xi'an Jiaotong University, Xi'an, China. His research interests include computer vision, pattern recognition, and machine learning. He is the author of more than 80 peer reviewed publications in prestigious
international journals and conferences. He is an area chair of WACV'2024\&2025, ICPR'2022\&2024, and CVPR'2022. He is an associate editor of PR, MVA, and PRL.
\end{IEEEbiography}

\begin{IEEEbiography}[{\includegraphics[width=1in,height=1.25in,clip,keepaspectratio]{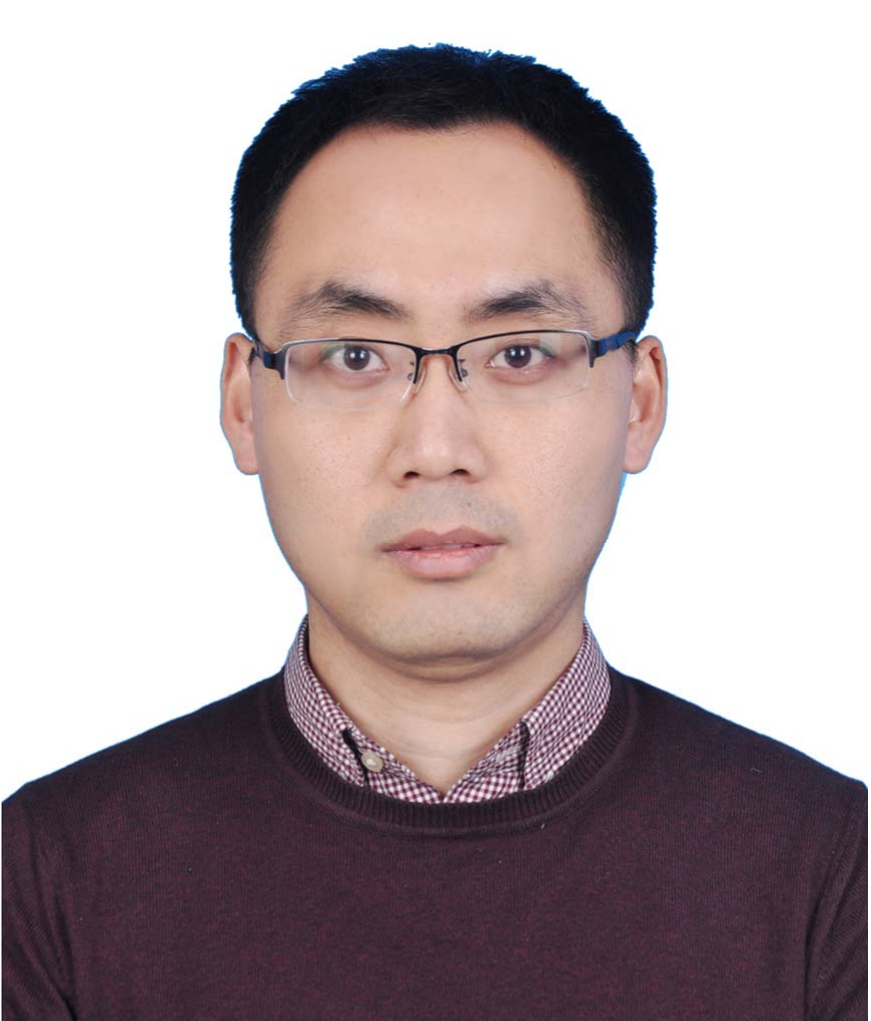}}]{Sanping Zhou} (Member, IEEE)
	received Ph.D. degree in control science and engineering from Xian Jiaotong University, Xi'an, China, in 2020. From 2018 to 2019, he was a visiting Ph.D. student at the Robotics Institute, Carnegie Mellon University. He is currently an Associate Professor with the Institute of Artificial Intelligence and Robotics, Xian Jiaotong University, Xi'an, China. His research interests include machine learning and computer vision, with a focus on object detection, image segmentation, visual tracking, multitask learning, and metalearning.
\end{IEEEbiography}

\begin{IEEEbiography}[{\includegraphics[width=1in,height=1.25in,clip,keepaspectratio]{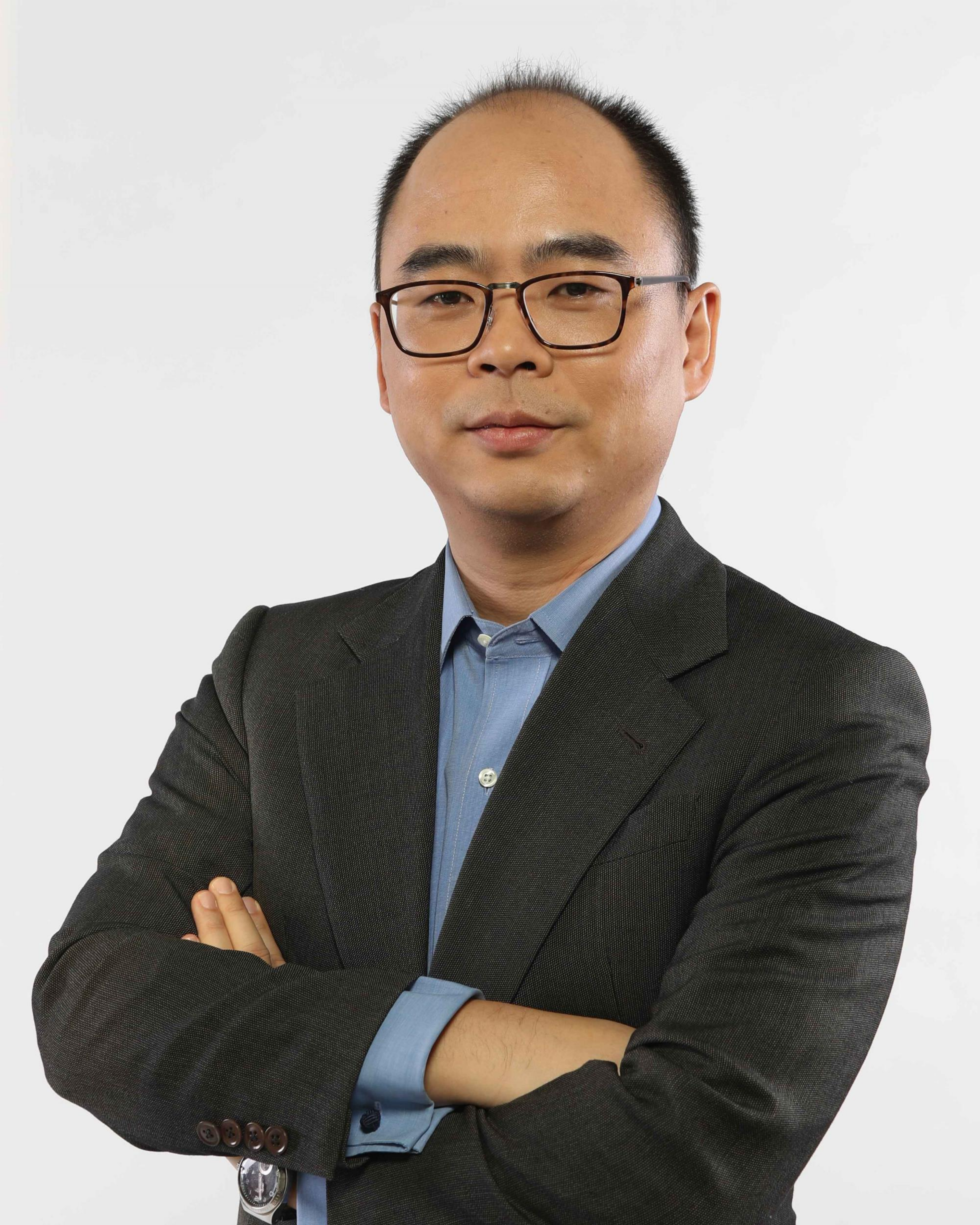}}]{Gang Hua} (Fellow, IEEE) received the B.S. and M.S. degrees in Automatic Control Engineering from Xi'an Jiaotong University (XJTU), Xi'an, China, in 1999 and 2002, respectively. He received the Ph.D. degree in Electrical Engineering and Computer Science at Northwestern University, Evanston, Illinois, USA, in 2006. He is currently the Vice President of the Multimodal Experiences Research Lab at Dolby Laboratories. His research focuses on computer vision, pattern recognition, machine learning, robotics, towards general Artificial Intelligence, with primary applications in cloud and edge intelligence. Before that, he was the CTO of Convenience Bee, and  the Managing Director and Chief Scientist of its research branch in US, Wormpex AI Research (2018-2024). He also served in various roles at Microsoft (2015-18) as the Science/Technical Adviser to the CVP of the Computer Vision Group, Director of Computer Vision Science Team in Redmond and Taipei ATL, and Senior Principal Researcher/Research Manager at Microsoft Research . He was an Associate Professor at Stevens Institute of Technology (2011-15). During 2014-15, he took an on leave and worked on the Amazon-Go project. He was a Visiting Researcher (2011-14) and a Research Staff Member (2010-11) at IBM Research T. J. Watson Center, a Senior Researcher (2009-10) at Nokia Research Center Hollywood, and a Senior Scientist (2006-09) at Microsoft Live labs Research. He is an associate editor of TPAMI and MVA. He is a general chair of ICCV'2027 and a program chair of CVPR'2019\&2022. He is the author of more than 200 peer reviewed publications in prestigious international journals and conferences. He holds 35 US patents and has 15 more US patents pending. He is the recipient of the 2015 IAPR Young Biometrics Investigator Award. He is an IEEE Fellow, an IAPR Fellow, and an ACM Distinguished Scientist.
\end{IEEEbiography}

\begin{IEEEbiography}[{\includegraphics[width=1in,height=1.25in,clip,keepaspectratio]{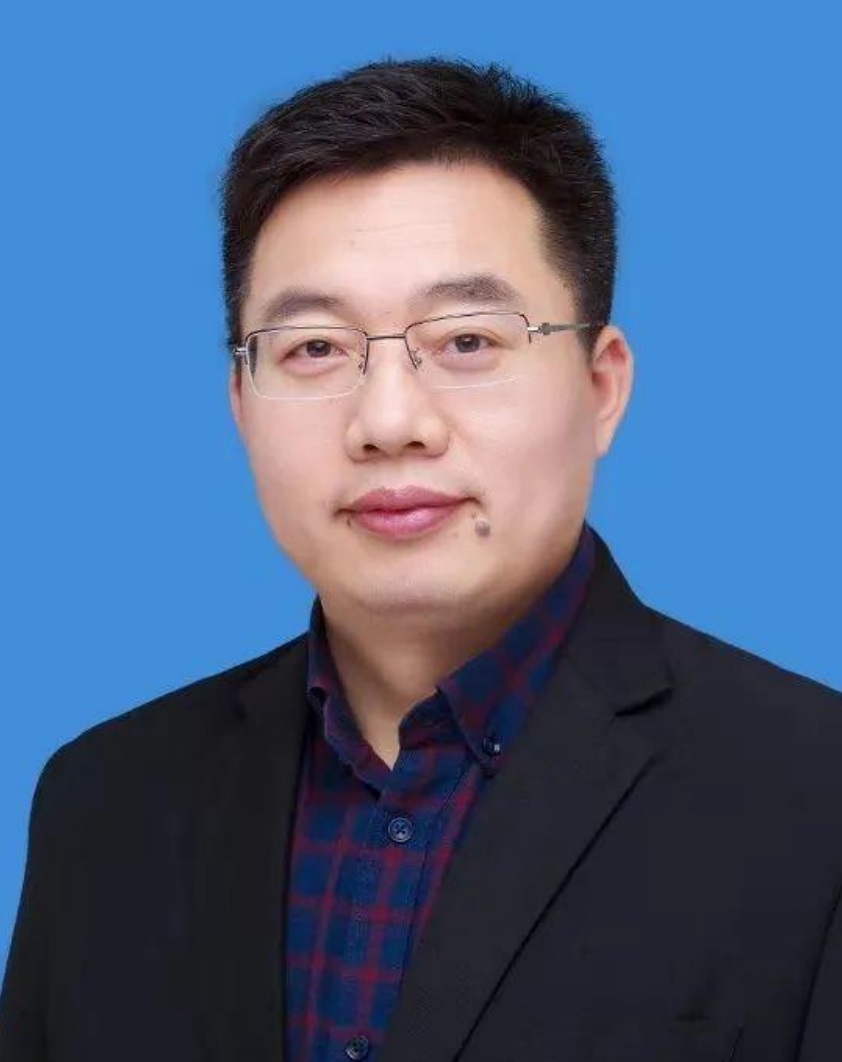}}]{Changyin Sun} 
	(Senior Member, IEEE) received the B.S. degree in Applied Mathematics from the College of Mathematics, Sichuan University, Chengdu, China, in 1996, the M.S. and Ph.D. degrees in Electrical Engineering from Southeast University, Nanjing, China, in 2001 and 2004, respectively. He is currently a Professor with the School of Artificial Intelligence, Anhui University, Hefei, China. His research interests include artificial intelligence, intelligent control, and optimal theory. He is an associate editor of TNNLS and IEEE/CAA JAS.
\end{IEEEbiography}




\end{document}